\documentclass[conference]{IEEEtran}
\IEEEoverridecommandlockouts

\usepackage{cite}
\usepackage{tikz}
\usepackage{amsmath,amssymb,amsfonts}
\usepackage{algorithmic}
\usepackage{hyperref}
\usepackage{booktabs}
\usepackage{bbm} 
\usepackage[font=footnotesize]{subfig}
\usepackage{amsmath,amssymb,amsfonts}
\usepackage{algorithmic}
\usepackage{graphicx}
\usepackage{textcomp}
\usepackage{mathtools}
\usepackage{amsthm} 
\usepackage{cleveref}
\usepackage{xcolor}
\def\BibTeX{{\rm B\kern-.05em{\sc i\kern-.025em b}\kern-.08em
    T\kern-.1667em\lower.7ex\hbox{E}\kern-.125emX}}

\newtheorem{theorem}{Theorem}[section]

\newcommand{\vecZ}{\mathbf{z}}

\newcommand{\matZ}{\mathbf{Z}}

\begin{document}

\title{DecoVAE: a Lightweight Interpretable Trend-Seasonal VAE Framework for Efficient Probabilistic Time Series Forecasting\\
}

\author{
\IEEEauthorblockN{1\textsuperscript{st} Alexander Marusov\IEEEauthorrefmark{1}}
\IEEEauthorblockA{
\textit{Applied AI Institute}\\
Moscow, Russia}
\and
\IEEEauthorblockN{1\textsuperscript{st} Dmitry Anikin\IEEEauthorrefmark{1}}
\IEEEauthorblockA{
\textit{Applied AI Institute}\\
Moscow, Russia}
\and
\IEEEauthorblockN{2\textsuperscript{nd} Alexey Zaytsev}
\IEEEauthorblockA{
\textit{Applied AI Institute}\\
Moscow, Russia\\
}

\thanks{\IEEEauthorrefmark{1}These authors contributed equally to this work.}
}

\maketitle

\begin{abstract}
Probabilistic time series forecasting remains challenging, largely because modeling distinct trend and seasonal dynamics requires specialized approaches. Existing methods often fail to capture the unique inner properties of these components, lack interpretability, or suffer from heavy memory and runtime overhead. To address these limitations, we propose DecoVAE, a lightweight interpretable trend-seasonal VAE framework that explicitly decomposes time series into trend and seasonal components by applying domain-specific inductive biases. The trend stream enforces structural smoothness using a 
differential regularizer on the latent trajectory, analogous to the Hodrick-Prescott filter. Concurrently, the seasonal stream operates in the frequency domain via a complex Gaussian VAE, natively capturing the amplitude and phase of periodic patterns. Extensive evaluations across seven real-world benchmarks show that DecoVAE consistently outperforms strong baselines. It achieves reductions of up to 14.96\% in CRPS and 23.30\% in NMAE for short-term forecasting, and up to 52.68\% and 26.51\% for long-term horizons. Crucially, DecoVAE yields these accuracy gains while remaining highly efficient, reducing model weight by up to 93\% and accelerating 
speed by up to 74\% compared to the second-best method. 
\end{abstract}

\begin{IEEEkeywords}
probabilistic time series forecasting, trend-seasonal decomposition, variational autoencoder, lightweight interpretable models
\end{IEEEkeywords}

\section{Introduction}

Probabilistic time series forecasting is essential for energy, finance, weather, and traffic applications~\cite{qiao2026s, kong2025deep, zhang2024probts}, yet remains challenging due to the need for accurate uncertainty quantification and effective modeling of trend (long-term, low-frequency movements) and seasonality (periodic oscillations)~\cite{PROIETTI202362}.

Current methods face several critical limitations: most treat trend and seasonality uniformly, ignoring their distinct structures; time-domain models struggle with global periodicity~\cite{zhang2025time}, while frequency-domain approaches suffer from spectral mixing that degrades both point forecasts and uncertainty estimates~\cite{an2026fredn, lu2026towards}. Although explicit decomposition improves accuracy~\cite{deng2024parsimony}, it intensifies the quality-computation trade-off, as high-fidelity generative models like diffusion networks remain prohibitively slow due to iterative sampling~\cite{10.1145/3783986, tashiro2021csdi, ALKILANE2026130269, rasul2021autoregressive}. Consequently, developing a computationally efficient framework that accurately captures both temporal dynamics for probabilistic forecasting remains an open challenge.

To address this gap, we introduce \textbf{DecoVAE (Trend-Seasonal Decomposition Variational Autoencoder)}, which explicitly separates time series into trend and seasonal components, modeling each via specialized mechanisms. We enforce trend smoothness through Hodrick--Prescott-inspired regularization, while modeling seasonality in the complex domain to naturally capture periodicity. Integrated into a variational framework, DecoVAE simultaneously captures structured temporal dynamics and provides well-calibrated probabilistic forecasts.

The main contributions of this work are summarized as follows:
\begin{itemize}
    \item \textbf{DecoVAE: a Lightweight Trend-Seasonal VAE Framework for Efficient Probabilistic Time Series Forecasting} 
    We propose DecoVAE, a variational autoencoder that performs explicit trend--seasonal decomposition with component-specific subnetworks, enabling accurate modeling of distinct temporal dynamics while yielding well-calibrated predictive distributions.
    
    \item \textbf{Theoretically Grounded Trend Regularization.} We enforce trend smoothness via a Hodrick--Prescott (HP)-inspired penalty~\cite{hodrick1997postwar}. Theorem~\ref{thm:hp_equivalent} shows equivalence to the classical HP objective under linear decoding, bridging statistical filtering with deep probabilistic modeling. Theorem~\ref{thm:nonlinear-decoder} further guarantees that, for Lipschitz decoders, the extracted trend exhibits negligible high-frequency energy, ensuring faithful long-term dynamic capture.
    
    \item \textbf{Complex-Domain Seasonal Modeling.} 
    We model seasonality in the complex frequency domain to natively encode phase and amplitude information. Inspired by FITS~\cite{xufits}, our complex-valued transformations overcome representational limits of real-valued time-domain processing, enhancing multi-periodic pattern learning within the variational framework.

    \item \textbf{Empirical validation.} Evaluations across seven benchmarks confirm DecoVAE's robustness across diverse forecasting scenarios. In short-term settings, it achieves up to 14.96\% CRPS and 23.30\% NMAE improvements over strong baselines; for long-term forecasts, gains reach 52.68\% CRPS and 26.51\% NMAE. Code is available at GitHub\footnote{\url{https://anonymous.4open.science/r/DecoVAE-3387/}}.
\end{itemize}

\section{Related Work}

This section reviews decomposition-based probabilistic forecasting methods by focusing on three key aspects: (A) probabilistic forecasting models, (B) techniques for decomposing and modeling trend and seasonal components (C), and loss functions that exploit decomposed temporal structure (D).

\subsection{Probabilistic Time Series Forecasting}
Generative models form the backbone of modern probabilistic time series forecasting. DeepAR \cite{salinas2020deepar} establishes a strong autoregressive baseline via global latent variables. VAE-based approaches capture uncertainty, with LaST \cite{wang2022learning} using a factorized latent space to disentangle components, and $K^2$VAE \cite{wu2025k2vae} integrating Koopman operator theory for long-term forecasting. Diffusion models like TSDiff-Cond \cite{kollovieh2023predict} offer flexible generation but at high computational cost. Alternatively, Trajectory Flow Matching (TFM) \cite{zhang2024trajectory} provides more efficient solution via continuous normalizing flows. Building on Neural ODE-based trajectory modeling, DeNOTS \cite{kuleshov2026denots} further improves stability and expressiveness while better preserving long-term temporal dependencies. Despite their accuracy, these methods typically trade off computational efficiency and interpretability, motivating a unified framework excelling across all three dimensions.

\subsection{Trend-Seasonal Decomposition Techniques}
Classical methods like STL~\cite{cleveland1990stl} and moving-average decomposition~\cite{Hyndman2011, kreuzer2025unpacking} separate trend (long-term direction) from seasonality (fixed-interval repetitions). Recent work enhances this: FedFormer~\cite{zhou2022fedformer} adaptively combines multi-scale moving averages, while Learnable Decomposition~\cite{yu2024revitalizing} replaces fixed kernels with trainable convolutions. Empirical studies confirm that explicit trend-seasonality separation consistently improves forecasting performance~\cite{luo2025decompnet}.

\subsection{Latent Modeling of Trend and Seasonality}
Modern architectures embed component-specific inductive biases. Autoformer~\cite{wu2021autoformer} applies asymmetric attention (complex for seasonality, simple for trend). FEDformer~\cite{zhou2022fedformer} operates in the frequency domain via sparse Fourier mode selection. Alternative designs include xPatch~\cite{stitsyuk2025xpatch} (separate MLP/convolution streams), TimeMixer (multi-scale pooling), TimesNet~\cite{wutimesnet} (2D period-aligned CNNs), and LaST~\cite{wang2022learning} (VAE with smoothness/periodicity constraints on latent components). These approaches balance expressiveness, efficiency, and interpretability while natively modeling distinct temporal structures.

\subsection{Loss Function}
Loss design critically affects decomposition-based forecasting. LaST~\cite{wang2022learning} applies auxiliary objectives to enforce smooth trend latents and periodic seasonal latents. DBLoss~\cite{dbloss2025} decomposes both predictions and targets via exponential smoothing, computing separate trend/seasonal losses. This explicit decomposition in the objective accelerates convergence and improves accuracy over standard MSE, even when the model internally disentangles components.



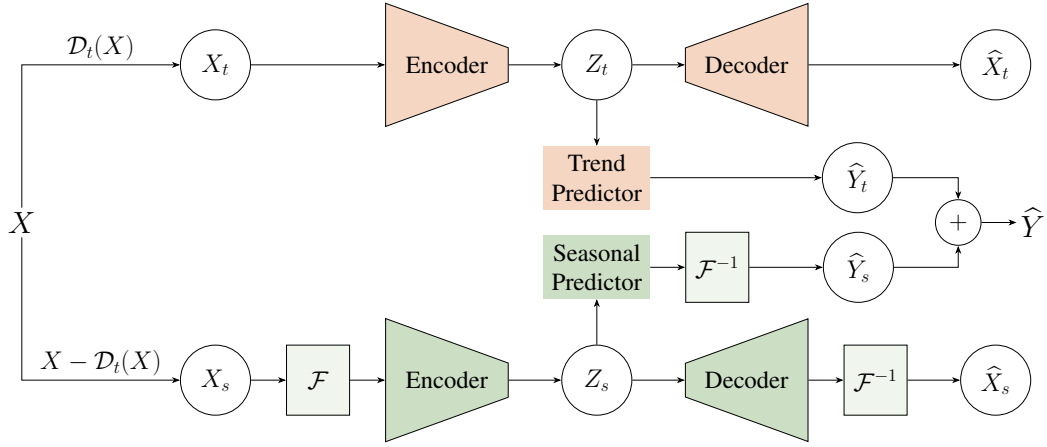
\begin{figure*}[h!]
    \centering
    \resizebox{0.8\textwidth}{!}{\usetikzlibrary{
    shapes.geometric, 
    arrows.meta,      
    positioning,      
    calc              
}

\definecolor{trendcolor}{HTML}{F4D6C2}
\definecolor{seasoncolor}{HTML}{CCDEC3}

\begin{tikzpicture}[
    node distance=1.0cm and 1.5cm,
    var/.style={font=\Huge},
    circ_node/.style={circle, draw, thick, minimum size=2cm, inner sep=0pt, font=\huge},
    sq_node/.style={rectangle, draw, thick, minimum size=2cm, font=\huge},
    sum_node/.style={circle, draw, thick, minimum size=1.3cm, font=\huge},
    pred_node/.style={rectangle, fill, text width=2.8cm, align=center, minimum height=1.8cm, font=\huge},
    enc_node/.style={trapezium, draw, thick, fill=#1, minimum width=1.8cm, minimum height=3.5cm, shape border rotate=270, trapezium angle=70, align=center, font=\huge},
    dec_node/.style={trapezium, draw, thick, fill=#1, minimum width=1.8cm, minimum height=3.5cm, shape border rotate=90, trapezium angle=70, align=center, font=\huge},
    fourier_node/.style={rectangle, draw, thick, fill=#1, minimum width=1.8cm, minimum height=2cm, align=center, font=\huge},
    arrow/.style={-Stealth, thick}
]


\node (Xs) [circ_node] {$X_{s}$};
\node (Fourier) [fourier_node=seasoncolor!30, right=1.0cm of Xs] {$\mathcal{F}$};
\node (Es) [enc_node=seasoncolor, right=1.0cm of Fourier] {Encoder};
\node (Zs) [circ_node, right=1.5cm of Es] {$Z_{s}$};
\node (Ds_dec) [dec_node=seasoncolor, right=1.5cm of Zs] {Decoder};
\node (InvFourier) [fourier_node=seasoncolor!30, right=1.0cm of Ds_dec] {$\mathcal{F}^{-1}$};
\node (Xhat_s) [circ_node, right=1.5cm of InvFourier] {$\widehat{X}_{s}$};

\node (Xt) [circ_node] at (Xs |- 0,9) {$X_{t}$};
\node (Et) [enc_node=trendcolor] at (Es |- 0,9) {Encoder};
\node (Zt) [circ_node] at (Zs |- 0,9) {$Z_{t}$};
\node (Dt_dec) [dec_node=trendcolor] at (Ds_dec |- 0,9) {Decoder};
\node (Xhat_t) [circ_node] at (Xhat_s |- 0,9) {$\widehat{X}_{t}$};

\node (Pred_t) [pred_node, fill=trendcolor, below=1.3cm of Zt] {Trend \\ Predictor};
\node (Pred_s) [pred_node, fill=seasoncolor, above=1.3cm of Zs] {Seasonal \\ Predictor};

\node (Yhat_t) [circ_node, right=4.9cm of Pred_t] {$\widehat{Y}_{t}$};

\node (InvFourier_pred)
    [fourier_node=seasoncolor!30, right=1.0cm of Pred_s]
    {$\mathcal{F}^{-1}$};

\node (Yhat_s)
    [circ_node, right=2.1cm of InvFourier_pred]
    {$\widehat{Y}_{s}$};

\coordinate (fork_point) at ($(Xt.west)!0.5!(Xs.west)$);
\node (X) [var, left=4.0cm of fork_point] {$X$};

\coordinate (mid_yhats) at ($(Yhat_t)!0.5!(Yhat_s)$);
\node (Sum) [sum_node, right=2.2cm of mid_yhats] {$+$};

\node (Yhat) [var, right=1cm of Sum] {$\widehat{Y}$};


\draw[arrow] (X) |- (Xs.west) node[pos=0.75, above, font=\huge] {$X -\mathcal{D}_t(X)$};
\draw[arrow] (X) |- (Xt.west) node[pos=0.75, above, font=\huge] {$\mathcal{D}_t(X)$};

\draw[arrow] (Xt.east) -- (Et.west);
\draw[arrow] (Et.east) -- (Zt.west);
\draw[arrow] (Zt.east) -- (Dt_dec.west);
\draw[arrow] (Dt_dec.east) -- (Xhat_t.west);

\draw[arrow] (Xs.east) -- (Fourier.west);
\draw[arrow] (Fourier.east) -- (Es.west);
\draw[arrow] (Es.east) -- (Zs.west);
\draw[arrow] (Zs.east) -- (Ds_dec.west);
\draw[arrow] (Ds_dec.east) -- (InvFourier.west);
\draw[arrow] (InvFourier.east) -- (Xhat_s.west);

\draw[arrow] (Zt.south) -- (Pred_t.north);
\draw[arrow] (Zs.north) -- (Pred_s.south);

\draw[arrow] (Pred_t.east) -- (Yhat_t.west);
\draw[arrow] (Pred_s.east) -- (InvFourier_pred.west);
\draw[arrow] (InvFourier_pred.east) -- (Yhat_s.west);

\draw[arrow] (Yhat_t.east) -| (Sum.north);
\draw[arrow] (Yhat_s.east) -| (Sum.south);
\draw[arrow] (Sum.east) -- (Yhat.west);

\end{tikzpicture}}
    \caption{The DecoVAE architecture decomposes the input time series $\mathbf{X}$ into trend ($\mathbf{X}_t$) and seasonal ($\mathbf{X}_s$) components. These are processed independently, accounting for trend smoothness and seasonal frequencies. The final prediction $\widehat{\mathbf{Y}}$ is synthesized as the sum of the forecasted trend $\widehat{\mathbf{Y}}_t$ and seasonality $\widehat{\mathbf{Y}}_s$.
    }
    \label{fig:decovae}
\end{figure*}

\section{Method}

\textbf{Problem statement.} 
To address the probabilistic time series forecasting problem, we consider a multivariate time series with $F$ features, where $\mathbf{x}_{\tau} \in \mathbb{R}^F$ represents the feature vector at timestamp $\tau$. The objective is to map historical observations $\mathbf{x}_{\tau-T+1:\tau}$ (lookback window $T$) to future values $\mathbf{x}_{\tau+1:\tau+L}$ (prediction horizon $L$). 


\textbf{Method pipeline.}
To tackle the probabilistic time series forecasting problem we propose DecoVAE (Fig.~\ref{fig:decovae}), a dual-stream Variational Autoencoder. It is designed to explicitly model the trend and seasonal components of a time series using structurally appropriate inductive biases:
\begin{itemize}
    \item a \textbf{smooth latent trajectory} for the trend;
    \item a \textbf{frequency-domain variational representation} for seasonality.
\end{itemize}

Thus, DecoVAE operates through the following pipeline:
\begin{enumerate}
    \item[A)] \textbf{Trend-Seasonal Decomposition Techniques}: The input series is decomposed into trend and seasonal components using moving average filtering.
    
    \item[B)] \textbf{Latent Modeling of Trend and Seasonality}: Each component is modeled via a dedicated VAE stream. Specifically, the trend component is processed directly within the time domain, whereas the seasonal component is mapped into the frequency domain using a Fourier transform.
    
    \item[C)] \textbf{Forecasting Pipeline}: Each stream generates independent predictions in its respective domain, which are combined to produce the final probabilistic forecast.

    \item[D)] \textbf{Loss Function}: For each component, we designed a specific objective that accounts for the concrete specifics:
    \begin{itemize}
        \item The \textit{trend stream} operates in the time domain with smoothness regularization applied to the latent trajectory, inspired by the Hodrick-Prescott filter.
        \item The \textit{seasonal stream} operates in the frequency domain using complex-valued representations to naturally capture amplitude and phase information of periodic patterns.
    \end{itemize}
\end{enumerate}

The remainder of this section details each stage of the pipeline. We first examine input decomposition strategies for isolating trend and seasonal components, followed by a description of the specialized processing pathways applied to each. Finally, we present the prediction mechanism and formalize the corresponding loss function.

\subsection{Trend-Seasonal Decomposition Techniques}

Given an input series $\mathbf{X} \in \mathbb{R}^{T \times F}$ with $T$ time steps and $F$ features, our goal is to  decompose it into trend and seasonal components via operator $\mathcal{D}_t$, which extracts trend component of the $\mathbf{X}$:
\begin{equation}
\mathbf{X} = \mathbf{X}_t + \mathbf{X}_s,
\end{equation}
where $\mathbf{X}_t = \mathcal{D}_t(\mathbf{X})$ and $\mathbf{X}_s = X -\mathcal{D}_t(\mathbf{X})$. Here, $\mathbf{X}_t, \mathbf{X}_s \in \mathbb{R}^{T \times F}$ denote the trend and seasonal components respectively.

We use a moving average (MA) trend extractor for its simplicity, with FEDFormer's MoE and EMA compared in Appendix~\ref{app:decomposition}. The separated components feed into specialized VAE pathways: the trend stream applies smoothness constraints to the latent trajectory, while the seasonal stream employs frequency-domain processing to capture periodicity.


\subsection{Latent Modeling of Trend and Seasonality}

We outline below our approach for modeling trend and seasonal components to yield their respective representations.

\subsubsection{\textbf{Trend Component}}

\paragraph{Trend Encoding}
The trend $\mathbf{X}_t$ is processed by a shared encoder applied independently along the channel dimension:
\begin{equation}
\mathbf{H} = \mathrm{MLP}^{enc}_{shared}(\mathbf{X}_t),
\quad
\mathbf{H} \in \mathbb{R}^{T \times d},
\end{equation}
where $T$ is the context length and $d$ is the embedding dimension. Following the standard VAE structure, we parameterize a Gaussian latent distribution. 
The mean and variance are computed from the encoder output:
\begin{align}
\boldsymbol{\mu} &= \mathrm{Linear}_\mu(\mathbf{H}) \in \mathbb{R}^{T \times d}, \\
\boldsymbol{\sigma}^2 &= \mathrm{Softplus}\big(\mathrm{Linear}_\sigma(\mathbf{H})\big) \in \mathbb{R}^{T \times d}.
\end{align}

\paragraph{Reparameterization and Sampling}
We use the standard reparameterization trick to obtain latent trend representations:
\begin{equation}
\mathbf{Z}_t = \boldsymbol{\mu} + \boldsymbol{\sigma} \odot \boldsymbol{\epsilon}, 
\quad \boldsymbol{\epsilon} \sim \mathcal{N}(0, I),
\end{equation}
where $\mathbf{Z}_t \in \mathbb{R}^{T \times d}$ is the latent trajectory, and $\odot$ denotes element-wise multiplication.

\paragraph{Decoding}
The latent trajectory is decoded to reconstruct the trend component. This is performed via MLP operating along the feature dimension (from embedding space to the original feature space):
\begin{equation}
\widehat{\mathbf{X}}_t = \mathrm{MLP}^{\mathrm{dec}}_{\mathrm{trend}}(\mathbf{Z}_t) \in \mathbb{R}^{T \times F}.
\end{equation}

\subsubsection{\textbf{Seasonal Component}}

To model seasonal dynamics, we operate directly in the frequency domain. 
This choice is motivated by the fact that periodic patterns correspond to localized structures in the spectrum, making them easier to represent and manipulate compared to the time domain~\cite{liu2024wftnet, crabbe2024time}.

We evaluated three alternative parameterizations for the frequency-domain seasonal component, ultimately selecting a \textit{complex-valued FITS-style approach}~\cite{xufits}. By modeling the signal directly within the complex Fourier domain via complex-valued linear layers, this method preserves the Fourier transform's native structure while jointly encoding amplitude and phase into a unified latent representation. As detailed in Appendix~\ref{app:seasonal_comparison}, this variant consistently outperforms alternatives, providing the most efficient spectral modeling and validating its adoption as our default.

\paragraph{Frequency Domain Encoding}
Let $\mathbf{X}_s \in \mathbb{R}^{T \times F}$ denote the seasonal component. After applying normalization, we transform seasonal component to the frequency domain:
\begin{equation}
\mathbf{X}^{(freq)}_s = \mathcal{F}(\mathbf{X}_s) \in \mathbb{C}^{K \times F},
\end{equation}
where $\mathcal{F}$ denotes the FFT and $K = \lfloor T/2 \rfloor + 1$ represents the number of frequency bins. The spectrum is encoded by a shared complex-valued linear encoder applied independently across frequency bins:
\begin{equation}
\mathbf{H} = \text{Linear}_{enc}(\mathbf{X}^{(freq)}_s) \in \mathbb{C}^{K \times d}.
\end{equation}

\paragraph{Complex Latent Distribution}
We parameterize a complex Gaussian distribution to preserve both amplitude and phase information. The mean is computed directly from the complex encoder output via complex linear layer. The variance, which must be real and positive, is computed from the concatenated real $\Re(\mathbf{H})$ and imaginary $\Im(\mathbf{H})$ parts:
\begin{align}
\boldsymbol{\mu} &= \text{Linear}_\mu(\mathbf{H}) \in \mathbb{C}^{K \times d}, \\
\boldsymbol{\sigma}^2 &= \text{Softplus}\left(\text{Linear}_\sigma([\Re(\mathbf{H}), \Im(\mathbf{H})])\right) \in \mathbb{R}^{K \times d}.
\end{align}

Sampling is performed via complex reparameterization:
\begin{equation}
\mathbf{Z}_s = \boldsymbol{\mu} + \sqrt{\frac{\boldsymbol{\sigma}^2}{2}} \odot (\boldsymbol{\epsilon}_r + i\,\boldsymbol{\epsilon}_i)
\end{equation}
where $\mathbf{Z}_{s} \in \mathbb{C}^{K \times d}$ and $\boldsymbol{\epsilon}_r, \boldsymbol{\epsilon}_i \sim \mathcal{N}(0, I)$. Symbol $\odot$ denotes element-wise multiplication.

\paragraph{Spectral Decoding}
The latent representation is decoded back to the frequency domain:
\begin{equation}
\widehat{\mathbf{X}}_s^{(freq)} = \text{Linear}_{dec}(\mathbf{Z}_s) \in \mathbb{C}^{K \times F}.
\end{equation}

\subsection{Forecasting Pipeline}

At inference time, the model generates probabilistic forecasts by sampling from each stream and combining their predictions.

\subsubsection{Trend Forecasting}
The latent trend trajectory $\mathbf{Z}_t$ is projected to the prediction horizon using two MLPs operating on temporal and feature dimensions:
\begin{align}
\widehat{\mathbf{Z}}_t &= \mathrm{MLP}^{pred}_1(\mathbf{Z}_t) \in \mathbb{R}^{L \times d} \\
\widehat{\mathbf{Y}}_t &= \mathrm{MLP}^{pred}_{2}(\widehat{\mathbf{Z}}_t) \in \mathbb{R}^{L \times F}.
\end{align}

Unlike the decoder which reconstructs the observed window, these prediction-specific MLPs learn temporal extrapolation from context length $T$ to forecast horizon $L$, accommodating potential distribution shifts between past and future.

\subsubsection{Seasonal Forecasting}
For forecasting a future window of length $L$, we use a separate predictor that upsamples the latent representation to $M = \lfloor L/2 \rfloor + 1$ frequency bins corresponding to the prediction horizon:
\begin{equation}
\widehat{\mathbf{Z}}_{s} = \text{Linear}^{pred}_1(\mathbf{Z}_s) \in \mathbb{C}^{M \times d}.
\end{equation}

This representation is decoded to the output channels:
\begin{equation}
\widehat{\mathbf{Y}}_s^{(freq)} = \text{Linear}^{pred}_2(\widehat{\mathbf{Z}}_{s}) \in \mathbb{C}^{M \times F}.
\end{equation}

The final seasonal forecast is obtained by inverse Fourier transform and and subsequent denormalization:
\begin{equation}
\widehat{Y}_{s} = \mathcal{F}^{-1}(\widehat{\mathbf{Y}}_s^{(freq)}) \in \mathbb{R}^{L \times F}.
\end{equation}

Such architecture enables efficient frequency-domain interpolation for forecasting, where the learned latent representation captures the essential seasonal patterns that can be extrapolated to future time horizons through spectral upsampling.

\subsubsection{Final Prediction}
The final forecast combines both components:
\begin{equation}
\widehat{\mathbf{Y}} = \widehat{\mathbf{Y}}_t + \widehat{\mathbf{Y}}_s, \quad \widehat{\mathbf{Y}}\in \mathbb{R}^{L \times F}.
\end{equation}
Probabilistic predictions are obtained by sampling multiple times from the latent distributions of both streams.

\subsection{Loss Function}

The model is trained by minimizing a composite loss function that combines the objectives for the trend and seasonal streams.

\subsubsection{Trend Loss}
To enforce smoothness of trend, we apply a differential regularizer on this latent trajectory, penalizing its first and second derivatives (velocity and acceleration). This approach is analogous to the Hodrick-Prescott filter applied in the latent space. The trend regularizer $\mathcal{R}_{trend}$ has the form:

\begin{equation}
\mathcal{R}_{trend} = \alpha_1 \mathcal{R}_1(\matZ_t) + \alpha_2 \mathcal{R}_2(\matZ_t)
\end{equation}
where $\mathcal{R}_1$ and $\mathcal{R}_2$ are the normalized squared norms of the first and second finite differences:
\begin{align}
\mathcal{R}_1(\matZ_t) &= \frac{1}{T-1} \sum_{\tau=1}^{T-1} \|\vecZ_{t,\tau+1}-\vecZ_{t,\tau}\|^2, \\
\mathcal{R}_2(\matZ_t) &= \frac{1}{T-2} \sum_{\tau=2}^{T-1} \|\vecZ_{t,\tau+1} - 2\vecZ_{t,\tau} + \vecZ_{t,\tau-1}\|^2.
\end{align}

The hyperparameters $\alpha_1 > 0$ and $\alpha_2 > 0$ control the contribution of velocity and curvature regularization, respectively. Both terms are necessary: the second derivative regularization allows piecewise smooth trends with potential jumps, while only first derivative regularization leads to linear trends. Their combination produces nonlinear but smooth trajectories appropriate for real-world time series.

Thus, the trend stream is trained by combining reconstruction error, KL divergence, and the smoothness regularizer:
\begin{equation}
\mathcal{L}_{trend} = \mathcal{L}_{rec}^{trend} + \mathcal{L}_{KL}^{trend} + \mathcal{R}_{trend},
\end{equation}
where:
\begin{itemize}
    \item $\mathcal{L}_{rec}^{trend} = \| \mathbf{X}_t - \widehat{\mathbf{X}}_t \|^2_2$ is the time-domain reconstruction loss.
    \item $\mathcal{L}_{KL}^{trend} = \mathrm{KL}\bigl(q_\phi(\mathbf{Z}_{t}) \,\|\, p(\mathbf{Z}_{t})\bigr)$ is the KL divergence for the trend latent variables.
    \item $\mathcal{R}_{trend}$ is the latent smoothness regularizer.
\end{itemize}

\begin{table*}[!htbp]
\centering
\caption{\textbf{Comparison of short-term probabilistic time series forecasting performance} ($T = 96$, $L=24$). Lower CRPS and NMAE scores indicate better performance. Results are reported as mean $\pm$ standard error, computed from \textbf{three independent runs} involving model retraining and evaluation. \textbf{Boldface} highlights the best-performing method, while \underline{underlining} denotes the second-best result.}
\label{tab:short_term}
\resizebox{\textwidth}{!}{
\begin{tabular}{llccccccc}
\toprule
Dataset & Metric & DecoVAE (ours) & $K^2$VAE & LaST & DeepAR & TFM & TSDiff & DeNOTS\\
\midrule

Electricity 
& CRPS & \underline{$0.088 \pm 0.001$} & $\mathbf{0.082 \pm 0.000}$ & $0.144 \pm 0.021$ & $0.117 \pm 0.001$ & $0.150 \pm 0.011$ & $0.115 \pm 0.006$ & $0.201 \pm 0.002$ \\
& NMAE & $\mathbf{0.104 \pm 0.001}$ & \underline{$0.107 \pm 0.000$} & $0.166 \pm 0.022$ & $0.151 \pm 0.002$ & $0.201 \pm 0.014$ & $0.150 \pm 0.008$ & $0.274 \pm 0.001$ \\

\midrule
Traffic 
& CRPS & \underline{$0.194 \pm 0.002$} & $\mathbf{0.164 \pm 0.001}$ & $0.309 \pm 0.011$ & $0.207 \pm 0.001$ & $0.321 \pm 0.013$ & $0.208 \pm 0.004$ & $0.386 \pm 0.019$ \\
& NMAE & \underline{$0.222 \pm 0.001$} & $\mathbf{0.211 \pm 0.002}$ & $0.360 \pm 0.010$ & $0.264 \pm 0.002$ & $0.426 \pm 0.022$ & $0.253 \pm 0.004$ & $0.500 \pm 0.032$ \\

\midrule
ETTh1 
& CRPS & $\mathbf{0.217 \pm 0.005}$ & \underline{$0.222 \pm 0.002$} & $0.254 \pm 0.004$ & $0.384 \pm 0.071$ & $0.540 \pm 0.018$ & $0.373 \pm 0.041$ & $0.503 \pm 0.008$ \\
& NMAE & $\mathbf{0.262 \pm 0.002}$ & \underline{$0.283 \pm 0.006$} & $0.289 \pm 0.005$ & $0.498 \pm 0.088$ & $0.623 \pm 0.026$ & $0.492 \pm 0.037$ & $0.645 \pm 0.017$ \\

\midrule
ETTh2 
& CRPS & $\mathbf{0.319 \pm 0.005}$ & \underline{$0.372 \pm 0.005$} & $0.460 \pm 0.058$ & $1.168 \pm 0.091$ & $1.738 \pm 0.451$ & $0.989 \pm 0.213$ & $1.920 \pm 0.398$ \\
& NMAE & $\mathbf{0.382 \pm 0.005}$ & \underline{$0.405 \pm 0.005$} & $0.546 \pm 0.077$ & $1.485 \pm 0.138$ & $1.892 \pm 0.384$ & $1.281 \pm 0.245$ & $2.207 \pm 0.421$ \\

\midrule
ETTm1 
& CRPS & \underline{$0.188 \pm 0.002$} & $\mathbf{0.186 \pm 0.005}$ & $0.217 \pm 0.004$ & $0.359 \pm 0.008$ & $0.556 \pm 0.038$ & $0.299 \pm 0.036$ & $0.561 \pm 0.048$ \\
& NMAE & $\mathbf{0.225 \pm 0.002}$ & \underline{$0.228 \pm 0.003$} & $0.257 \pm 0.008$ & $0.473 \pm 0.005$ & $0.634 \pm 0.038$ & $0.396 \pm 0.059$ & $0.721 \pm 0.050$ \\

\midrule
ETTm2 
& CRPS & $\mathbf{0.233 \pm 0.001}$ & $0.314 \pm 0.027$ & \underline{$0.274 \pm 0.004$} & $0.999 \pm 0.182$ & $1.784 \pm 0.121$ & $0.468 \pm 0.066$ & $1.591 \pm 0.889$ \\
& NMAE & $\mathbf{0.286 \pm 0.001}$ & \underline{$0.316 \pm 0.013$} & $0.325 \pm 0.007$ & $1.273 \pm 0.201$ & $1.904 \pm 0.151$ & $0.630 \pm 0.087$ & $1.815 \pm 0.906$ \\

\midrule
Weather 
& CRPS & $\mathbf{0.286 \pm 0.003}$ & $0.683 \pm 0.011$ & $0.538 \pm 0.057$ & $0.360 \pm 0.025$ & $1.008 \pm 0.322$ & \underline{$0.327 \pm 0.060$} & $0.434 \pm 0.022$ \\
& NMAE & $\mathbf{0.316 \pm 0.001}$ & $0.491 \pm 0.014$ & $0.676 \pm 0.090$ & $0.425 \pm 0.017$ & $1.206 \pm 0.403$ & \underline{$0.412 \pm 0.093$} & $0.562 \pm 0.030$ \\

\midrule
Avg. Rank
& CRPS & $\mathbf{1.429}$ & \underline{$2.286$} & $3.714$	& $4.286$ & $6.429$ & $3.571$ & $6.286$ \\
& NMAE & $\mathbf{1.143}$ & \underline{$2.143$} & $4.000$ & $4.429$ & $6.286$ & $3.429$ & $6.571$ \\

\bottomrule
\end{tabular}
}
\end{table*}

The proposed trend regularization has theoretical properties. Specifically, Theorem~\ref{thm:hp_equivalent} establishes that, under a linear trend decoder, our objective function closely approximates the classical Hodrick–Prescott loss.

\begin{theorem}[Hodrick--Prescott Filter Approximation in Latent Space]
\label{thm:hp_equivalent}
Assume the trend decoder is linear in the temporal domain, i.e., $\mathcal{G}_t(\mathbf{z}_{t,\tau}) = W_t \mathbf{z}_{t,\tau}$ with $W_t \in \mathbb{R}^{F \times d}$, and let $\alpha_1 = 0$. Define the latent objective functional:
\begin{equation}
J[Z_t] := \sum_{\tau=1}^T \|X_{t,\tau} - W_t z_{t,\tau}\|^2 + \alpha_2 \sum_{\tau=2}^{T-1} \|\Delta^2 z_{t,\tau}\|^2.
\end{equation}
Then, for any latent sequence $Z_t$, the discrepancy between the classical Hodrick--Prescott functional $J_{\mathrm{HP}}[W_t Z_t]$ and $J[Z_t]$ satisfies the two-sided bound:
\begin{equation}
\begin{aligned}
&\bigl(\lambda_{\mathrm{HP}}\sigma^2_{\min}(W_t) - \alpha_2\bigr) \sum_{\tau=2}^{T-1} \|\Delta^2 z_{t,\tau}\|^2 \\ 
& \quad \leq J_{\mathrm{HP}}[W_t Z_t] - J[Z_t] \\
& \quad \leq \alpha_2 \left( \frac{\sigma^2_{\max}(W_t)}{\sigma^2_{\min}(W_t)} - 1 \right) \sum_{\tau=2}^{T-1} \|\Delta^2 z_{t,\tau}\|^2,
\end{aligned}
\end{equation}
where $\sigma_{\min}(W_t)$ and $\sigma_{\max}(W_t)$ denote the smallest and largest singular values of $W_t$, respectively. By selecting $\lambda_{\mathrm{HP}} = \alpha_2 / \sigma^2_{\min}(W_t)$, the lower bound vanishes, and the functional difference is controlled solely by the condition number of the decoder matrix.
\end{theorem}

The complete proof is provided in the supplementary repository\footnote{\url{https://anonymous.4open.science/r/DecoVAE-3387/docs/Theorem1.pdf}}.

Building on this result, Theorem~\ref{thm:nonlinear-decoder} demonstrates that even under a nonlinear Lipschitz-continuous decoder, the reconstructed trend retains exclusively low-frequency spectral content. This establishes a formal guarantee of spectral disentanglement, ensuring that the trend component remains strictly smooth and free from high-frequency seasonal artifacts.

\begin{theorem}[Preservation of Spectral Separation under Nonlinear Decoding]
\label{thm:nonlinear-decoder}
Let the trend decoder $\mathcal{G}_t: \mathbb{R}^d \to \mathbb{R}^F$ be $L_t$-Lipschitz continuous in the temporal domain, i.e.,
\begin{equation}
\|\mathcal{G}_t(\mathbf{z}_{t,\tau_1}) - \mathcal{G}_t(\mathbf{z}_{t,\tau_2})\| \leq L_t \|\mathbf{z}_{t,\tau_1} - \mathbf{z}_{t,\tau_2}\|, \quad \forall \tau_1, \tau_2.
\end{equation}
Then, the spectral energy of the reconstructed trend in the high-frequency band $[\omega_h, \pi]$ satisfies:
\begin{equation}
\begin{aligned}
&\int_{\omega_h}^{\pi} \|\mathcal{F}(\mathcal{G}_t \circ \mathbf{z}_t)(\omega)\|^2 \, d\omega \\ 
& \quad \leq 2L_t^2 \int_{\omega_h/2}^{\pi} \|\mathcal{F}(\mathbf{z}_t)(\omega)\|^2 \, d\omega + \frac{8L_t^2 \|\mathbf{z}_t\|_{\ell^2}^2}{T \omega_h^2}.
\end{aligned}
\end{equation}
\end{theorem}

The proof can be found in the supplementary repository\footnote{\url{https://anonymous.4open.science/r/DecoVAE-3387/docs/Theorem2.pdf}}.


\subsubsection{Seasonal Loss}
In the final model configuration, no explicit spectral regularization is employed for the seasonal module. However, for completeness and symmetry with the trend module, we additionally explored several alternative regularization schemes, which are reported in the Appendix~\ref{app:season_variants}.

Aside from the regularization term, the seasonal loss is formulated analogously to the trend loss. The key distinction lies in the reconstruction term, which is applied in the frequency domain to emphasize spectral fidelity:
\begin{equation}
\mathcal{L}_{rec}^{season} = \left\| \widehat{\mathbf{X}}_s^{(freq)} - \widetilde{\mathbf{X}}_s \right\|_2^2 = \sum_{k=1}^K \sum_{f=1}^F |\widehat{X}_{s,k,f} - \widetilde{X}_{s,k,f}|^2.
\end{equation}

Thus, the seasonal stream is trained with a multi-component loss in the frequency domain:
\begin{equation}
\mathcal{L}_{season} = \mathcal{L}_{rec}^{freq} + \mathcal{L}_{KL}^{season},
\end{equation}
where:
\begin{itemize}
    \item $\mathcal{L}_{rec}^{freq}$ is the spectral reconstruction loss.
    \item $\mathcal{L}_{KL}^{season} = \mathrm{KL}\bigl(q_\psi(\mathbf{Z}_{s}) \,\|\, p(\mathbf{Z}_{s})\bigr)$ is the KL divergence for the seasonal latent variables.
\end{itemize}

\subsubsection{Final Loss Functional}
The total loss for DecoVAE is:
\begin{equation}
\label{eq:total-loss}
\mathcal{L}_{\text{total}} = \mathcal{L}_{\text{pred}} + \mathcal{L}_{\text{trend}} + \mathcal{L}_{\text{season}},
\end{equation}
where the prediction loss is:
\begin{equation}
\mathcal{L}_{\text{pred}} = \| \mathbf{Y} - (\widehat{\mathbf{Y}}_t + \widehat{\mathbf{Y}}_s)\|^2_2.
\end{equation}

\section{Experiments}

\begin{table*}[!htbp]
\centering
\caption{\textbf{Comparison of long-term probabilistic time series forecasting performance} ($T = 96$, $L=720$). Lower CRPS and NMAE scores indicate better performance. Results are reported as mean $\pm$ standard error, computed from \textbf{three independent runs} involving model retraining and evaluation. \textbf{Boldface} highlights the best-performing method, while \underline{underlining} denotes the second-best result.}
\label{tab:long_term}
\resizebox{\textwidth}{!}{
\begin{tabular}{llcccccccc}
\toprule
Dataset & Metric & DecoVAE (ours) & $K^2$VAE & LaST & DeepAR & TFM & TSDiff & DeNOTS\\
\midrule

Electricity 
& CRPS & \underline{$0.120 \pm 0.000$} & $\mathbf{0.115 \pm 0.006}$ & $0.146 \pm 0.001$ & $0.163 \pm 0.009$ & $0.210 \pm 0.017$ & $0.126 \pm 0.003$ & $0.186 \pm 0.009$ \\
& NMAE & $\mathbf{0.138 \pm 0.001}$ & \underline{$0.140 \pm 0.004$} & $0.169 \pm 0.001$ & $0.204 \pm 0.011$ & $0.292 \pm 0.022$ & $0.165 \pm 0.004$ & $0.186 \pm 0.009$ \\

\midrule
Traffic 
& CRPS & \underline{$0.218 \pm 0.001$} & $\mathbf{0.189 \pm 0.002}$ & $0.301 \pm 0.003$ & $0.220 \pm 0.003$ & $0.473 \pm 0.018$ & $0.297 \pm 0.027$ & $0.414 \pm 0.143$ \\
& NMAE & \underline{$0.246 \pm 0.002$} & $\mathbf{0.243 \pm 0.000}$ & $0.343 \pm 0.002$ & $0.286 \pm 0.006$ & $0.637 \pm 0.023$ & $0.368 \pm 0.033$ & $0.419 \pm 0.151$ \\

\midrule
ETTh1 
& CRPS & $\mathbf{0.298 \pm 0.003}$ & \underline{$0.334 \pm 0.008$} & $0.573 \pm 0.009$ & $0.690 \pm 0.211$ & $0.590 \pm 0.060$ & $0.575 \pm 0.059$ & $0.555 \pm 0.014$ \\
& NMAE & $\mathbf{0.360 \pm 0.004}$ & \underline{$0.428 \pm 0.012$} & $0.614 \pm 0.010$ & $0.830 \pm 0.240$ & $0.839 \pm 0.088$ & $0.701 \pm 0.054$ & $0.750 \pm 0.019$ \\

\midrule
ETTh2 
& CRPS & $\mathbf{0.478 \pm 0.008}$ & \underline{$1.000 \pm 0.235$} & $2.531 \pm 0.054$ & $2.332 \pm 0.013$ & $2.495 \pm 0.688$ & $2.254 \pm 0.106$ & $2.082 \pm 0.136$ \\
& NMAE & $\mathbf{0.580 \pm 0.001}$ & \underline{$0.769 \pm 0.083$} & $2.602 \pm 0.038$ & $2.611 \pm 0.053$ & $3.204 \pm 0.992$ & $2.488 \pm 0.104$ & $2.356 \pm 0.104$ \\

\midrule
ETTm1 
& CRPS & \underline{$0.293 \pm 0.001$} & $\mathbf{0.266 \pm 0.024}$ & $0.421 \pm 0.013$ & $0.466 \pm 0.073$ & $0.592 \pm 0.010$ & $0.483 \pm 0.020$ & $0.576 \pm 0.009$ \\
& NMAE & $\mathbf{0.327 \pm 0.002}$ & \underline{$0.328 \pm 0.039$} & $0.487 \pm 0.015$ & $0.623 \pm 0.069$ & $0.855 \pm 0.025$ & $0.629 \pm 0.025$ & $0.614 \pm 0.010$ \\

\midrule
ETTm2 
& CRPS & $\mathbf{0.468 \pm 0.005}$ & \underline{$0.989 \pm 0.275$} & $1.871 \pm 0.241$ & $2.064 \pm 0.646$ & $2.899 \pm 0.782$ & $1.642 \pm 0.450$ & $1.931 \pm 0.016$ \\
& NMAE & $\mathbf{0.524 \pm 0.001}$ & \underline{$0.713 \pm 0.105$} & $2.022 \pm 0.251$ & $2.388 \pm 0.633$ & $3.783 \pm 0.975$ & $1.915 \pm 0.430$ & $2.008 \pm 0.036$ \\

\midrule
Weather 
& CRPS & \underline{$0.431 \pm 0.002$} & $0.661 \pm 0.007$ & $0.672 \pm 0.028$ & $0.465 \pm 0.064$ & $1.344 \pm 0.246$ & $\mathbf{0.375 \pm 0.030}$ & $0.803 \pm 0.071$ \\
& NMAE & $\mathbf{0.459 \pm 0.005}$ & $0.606 \pm 0.008$ & $0.835 \pm 0.043$ & $0.555 \pm 0.067$ & $1.542 \pm 0.422$ & \underline{$0.465 \pm 0.033$} & $0.865 \pm 0.084$ \\

\midrule
Avg. Rank
& CRPS & $\mathbf{1.571}$ & \underline{$1.857$} & $4.571$ & $4.714$ & $6.714$ & $3.571$  & $5.000$ \\
& NMAE & $\mathbf{1.143}$ & \underline{$2.143$} & $4.143$ &  $5.000$ & $7.000$ & $3.857$ & $4.714$ \\

\bottomrule
\end{tabular}
}
\end{table*}

\subsection{Experiments Setup}

\paragraph{Datasets.} We evaluate the prediction quality using seven real-world datasets from diverse application domains. These include four variants of the Electricity Transformer Temperature dataset (ETTh1, ETTh2, ETTm1, ETTm2)~\cite{zhou2021informer}, traffic flow measurements~\cite{lai2018modeling}, household electricity consumption\footnote{\url{https://archive.ics.uci.edu/dataset/321/electricityloaddiagrams20112014}, accessed in May 2026.}, and meteorological observations\footnote{\url{https://www.bgc-jena.mpg.de/wetter/}, accessed in May 2026.}. Complete dataset specifications are provided in Appendix~\ref{app:datasets}.

\paragraph{Evaluation Protocol.}
Model performance is quantified using two complementary metrics: the Continuous Ranked Probability Score (CRPS)~\cite{gneiting2007strictly,zamo2018estimation}, which assesses distributional forecast quality, and Normalized Mean Absolute Error (NMAE), which evaluates point prediction accuracy. We refer readers to Appendix~\ref{app:metrics} for complete metric definitions.

\paragraph{Baselines.} We compare our approach against several generative models, including modern $K^2$VAE~\cite{wu2025k2vae}, LaST~\cite{wang2022learning}, TSDiff~\cite{kollovieh2023predict}, DeepAR~\cite{salinas2020deepar}, DeNOTS~\cite{kuleshov2026denots} and TFM~\cite{zhang2024trajectory}.

\subsection{Main Results}

\paragraph{Short-term forecasting ($T=96,L=24$)} 
For short-term forecasting (Table~\ref{tab:short_term}), DecoVAE yields substantial CRPS gains on ETTh2 (+14.25\%), ETTm2 (+14.96\%), and Weather (+12.54\% over TSDiff). On high-dimensional data, it secures the best NMAE on Electricity ($0.104 \pm 0.001$) and outperforms most baselines by over 30\% on Traffic. Overall, DecoVAE surpasses non-decomposition methods by over 30\% in CRPS, significantly outperforming LaST on ETTh2 (+30.65\%) and Weather (+46.84\%), while matching or exceeding $K^2$VAE despite a simpler architecture. Finally, DecoVAE achieves the best average ranks: 1.429 and 1.143 for CRPS and NMAE correspondingly surpassing all previous approaches.

\paragraph{Long-term forecasting ($T=96,L=720$)} 
For long-term forecasting (Table~\ref{tab:long_term}), DecoVAE's explicit decomposition yields substantial advantages. On ETT datasets, it improves CRPS by up to 52.68\% and NMAE by up to 26.51\% over second-best methods. Against LaST, it achieves up to 81.11\% CRPS improvement on ETTh2, highlighting the limitations of latent space decomposition over extended horizons, while vastly outperforming non-decomposition baselines like DeepAR. Additionally, it matches $K^2$VAE's NMAE ($0.138 \pm 0.001$) on Electricity with greater efficiency, and secures the best NMAE ($0.459 \pm 0.005$) on the Weather dataset.





\subsection{Computational Efficiency Analysis}

Beyond predictive accuracy, we evaluate DecoVAE's computational characteristics across training speed, memory consumption, model size, and inference time. We focus on the two largest datasets--Electricity (321 channels) and Traffic (862 channels)--considering both short-term ($L=24$) and long-term ($L=720$) horizons, comparing against K$^2$VAE and TSDiff, the best and second-best performing methods from our main experiments.

\paragraph{Time Efficiency}

\begin{figure}[!htbp]
    \centering
    \includegraphics[width=0.85\linewidth]{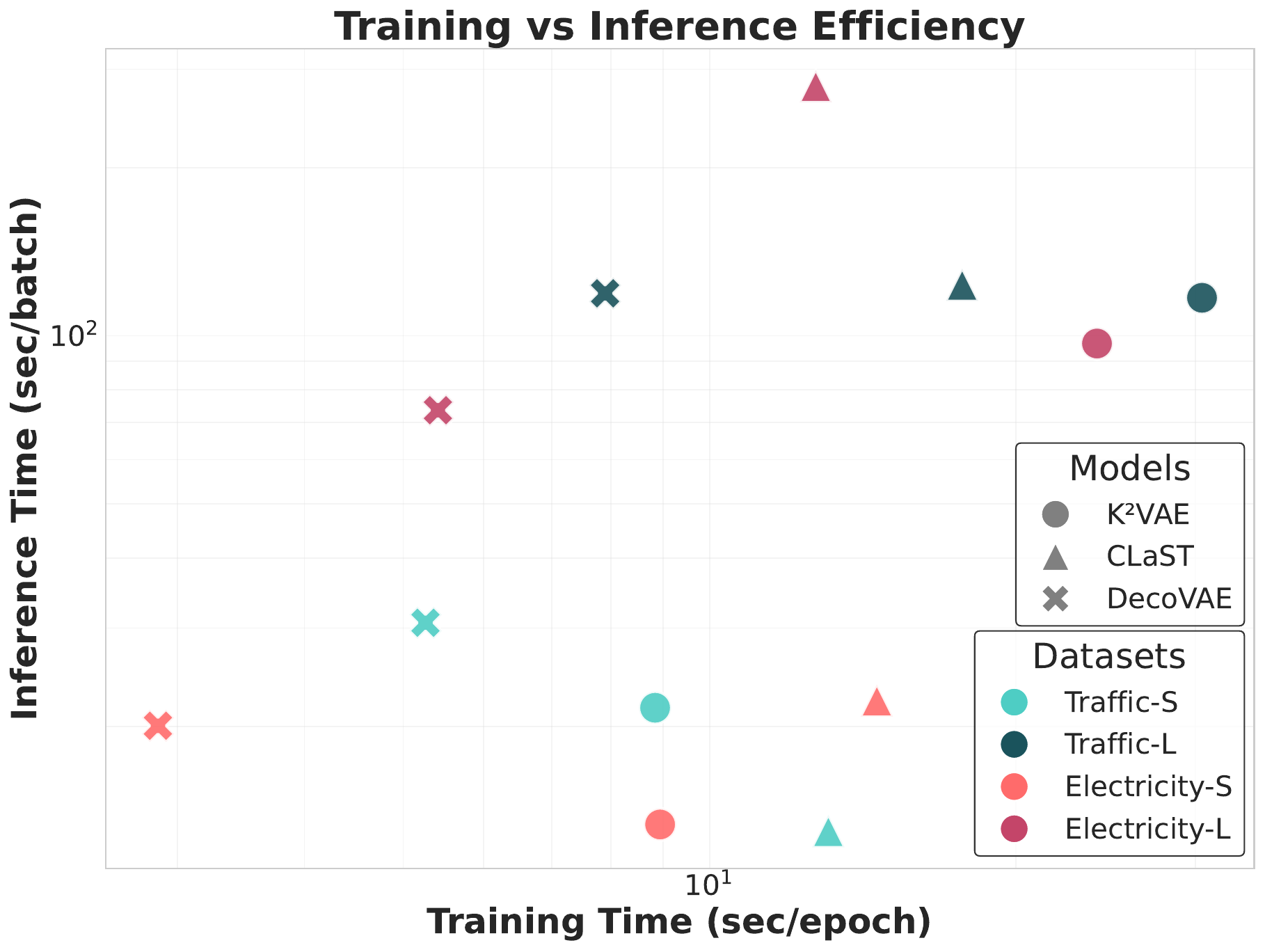}
    \caption{Comparison of training and inference time across models for short-term ($L=24$) and long-term ($L=720$) forecasting tasks on the Electricity and Traffic datasets. Lower values indicate better computational efficiency.}
    \label{fig:time_comparison}
\end{figure}

As shown in Figure~\ref{fig:time_comparison}, DecoVAE demonstrates substantial training efficiency advantages: 30--74\% faster per epoch compared to K$^2$VAE, with inference speed improvements of 10--35\%. Against TSDiff, DecoVAE achieves approximately 50\% faster training and 32--76\% faster inference, with greater gains in long-term forecasting.

\paragraph{Memory Consumption and Model Size}

\begin{figure}[!h]
    \centering
    \includegraphics[width=0.85\linewidth]{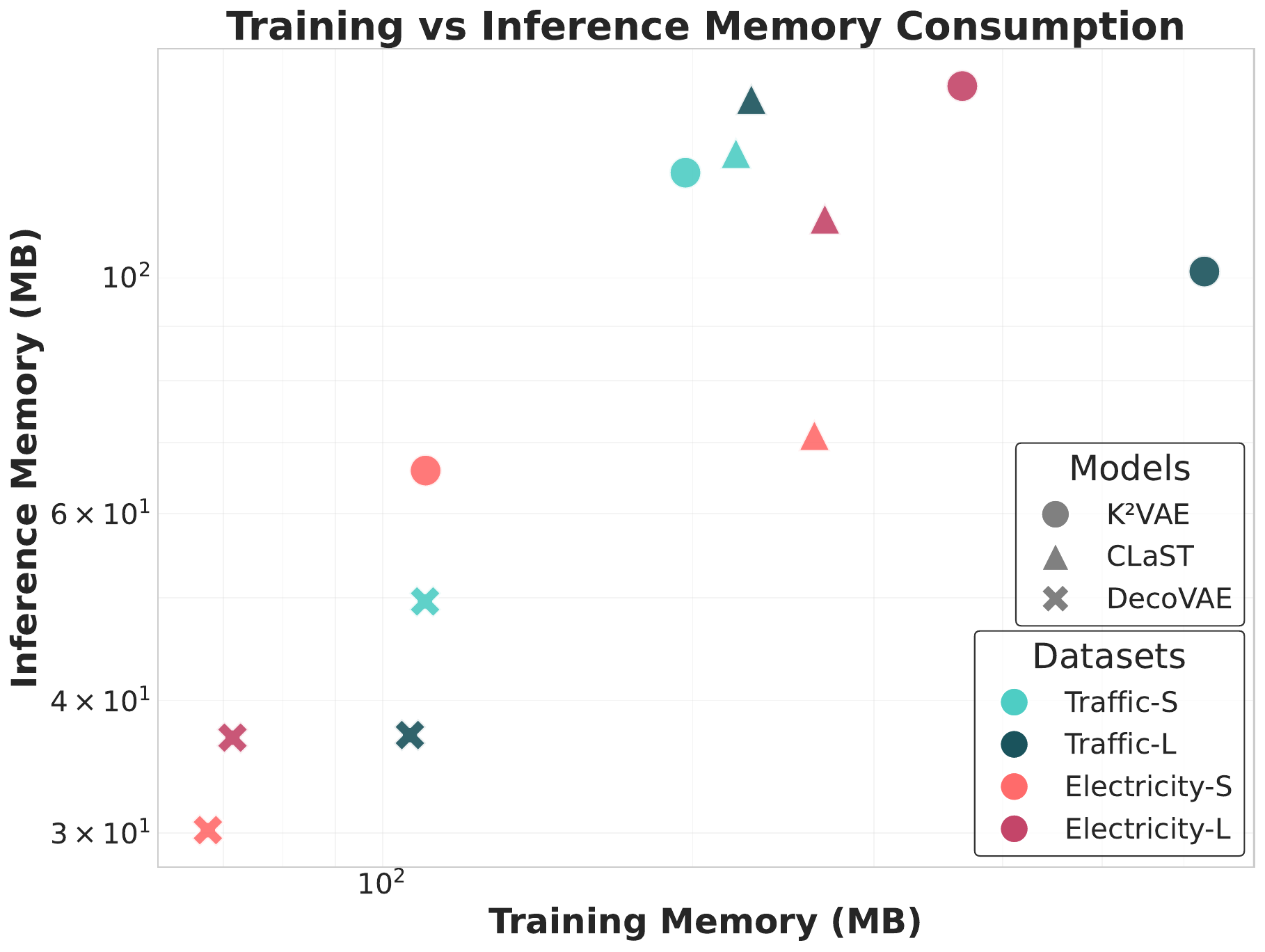}
    \caption{Comparison of training and inference memory consumption for short-term ($L=24$) and long-term ($L=720$) forecasting tasks on the Electricity and Traffic datasets. Lower values indicate better computational efficiency.}
    \label{fig:memory_comparison}
\end{figure}

Figure~\ref{fig:memory_comparison} shows DecoVAE achieves 48--85\% reduction in test memory usage and 80--93\% smaller model size compared to K$^2$VAE (5MB vs 70MB in extreme cases), corresponding to 85--95\% fewer parameters (under 1M vs up to 18M+). While TSDiff has 2--3$\times$ smaller model size than DecoVAE, this comes at the cost of significantly slower inference due to its iterative diffusion mechanism. DecoVAE remains 10--20$\times$ more compact than K$^2$VAE while delivering comparable forecasting performance. These results establish DecoVAE as computationally efficient, achieving favorable trade-offs between operational requirements and forecasting quality.

\subsection{Interpretability}

\begin{figure}[!h]
    \centering
    \includegraphics[width=1\linewidth]{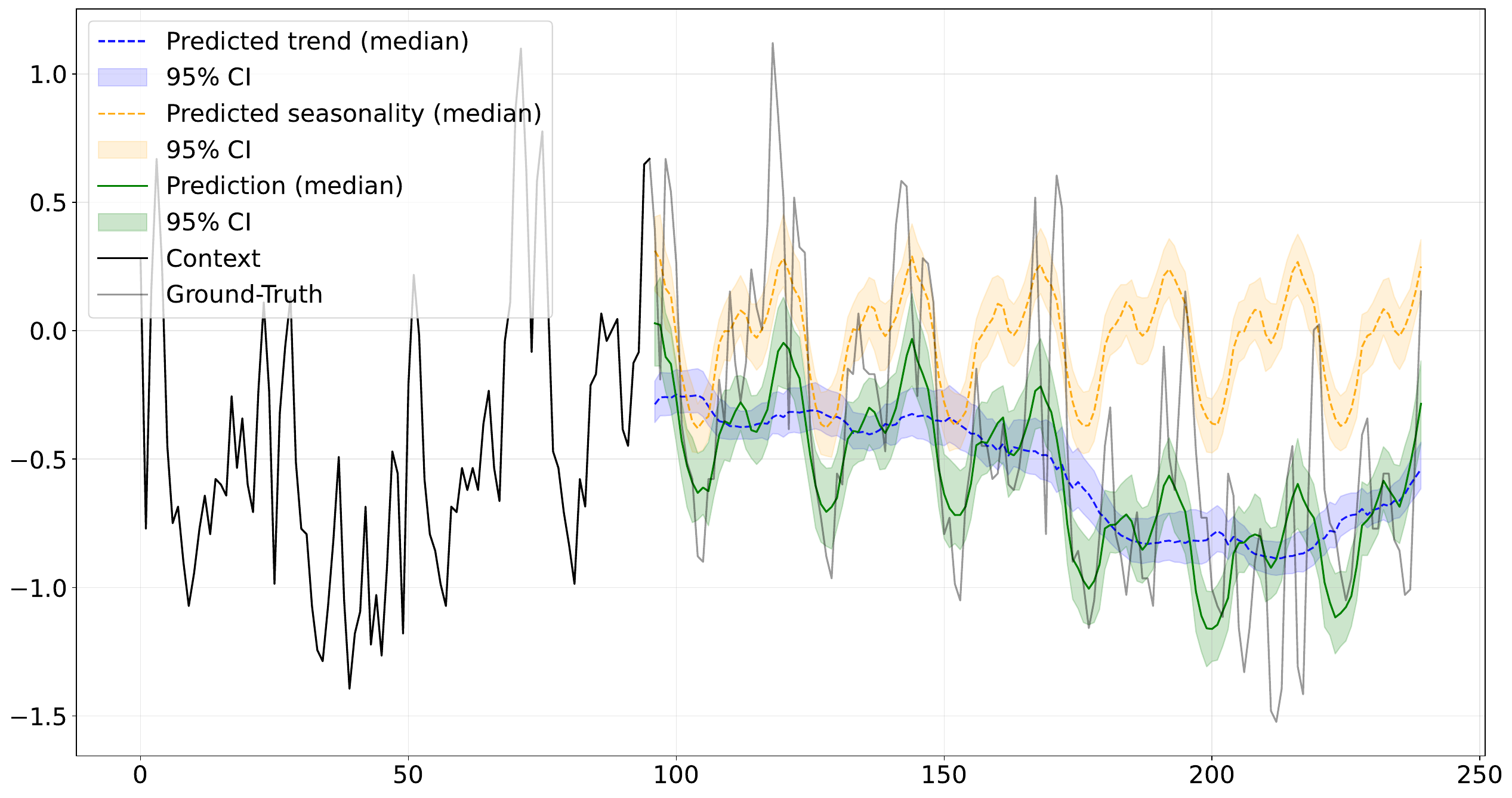}
    \caption{Probabilistic trend-seasonal forecasts generated by DecoVAE on the Electricity dataset, with ground-truth observations overlaid for comparison.}
    \label{fig:interpretability}
\end{figure}

DecoVAE provides interpretable predictions through explicit trend-seasonal decomposition. Figure$~\ref{fig:interpretability}$ illustrates a representative example on the Electricity dataset, where the model successfully disentangles long-term directional movement (trend, blue) from short-term periodic fluctuations (seasonality, yellow). DecoVAE achieves interpretable decomposition, preserving the input's structural and temporal dynamics. 

\begin{figure}[!h]
    \centering
    \includegraphics[width=0.85\linewidth]{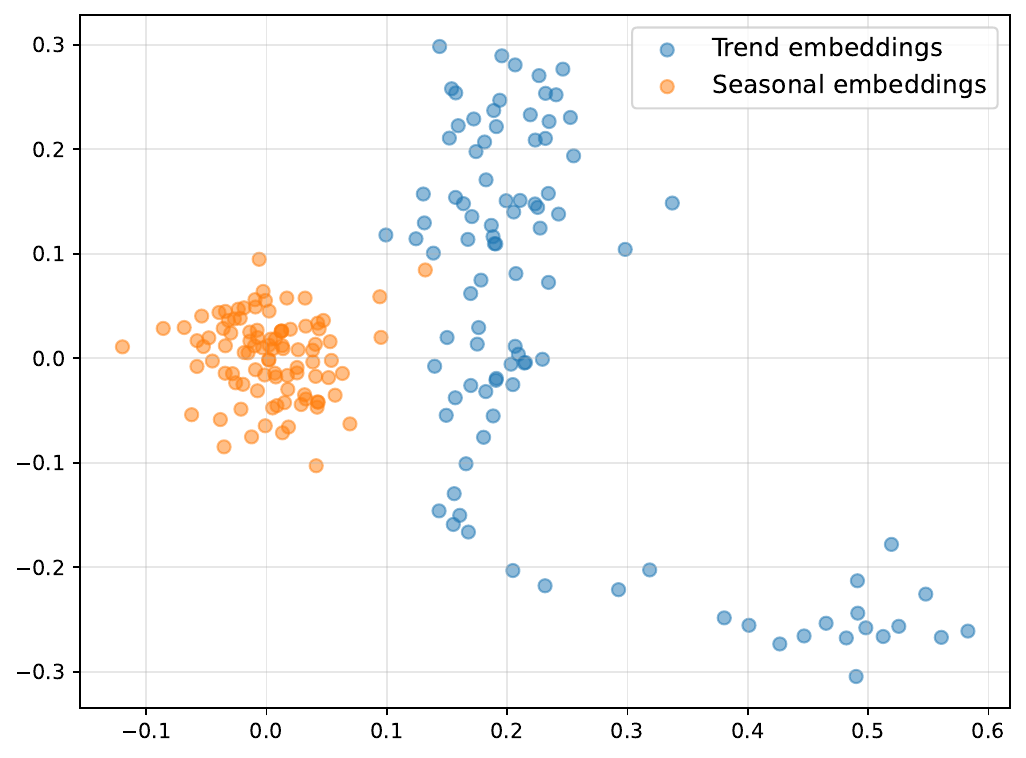}
    \caption{Principal component analysis (PCA) of trend and seasonal embeddings projected onto a 2D space on the Electricity dataset.}
    \label{fig:embeddings}
\end{figure}

Figure~\ref{fig:embeddings} presents a Principle Component Analysis (PCA) of trend and seasonal embeddings, with the seasonal embeddings first converted from the frequency domain to the time domain. Crucially, the two groups are highly separable. Trend embeddings form an almost distinct path, likely due to first- and second-derivative regularization, whereas seasonal embeddings appear more clustered. This clustering may indicate that, within a given periodicity, seasonal embeddings corresponding to similar periods are closely grouped, or as a result of the frequency-to-time domain transformation.

\subsection{Regularization Effect}

\begin{figure}
    \centering
    \includegraphics[width=0.9\linewidth]{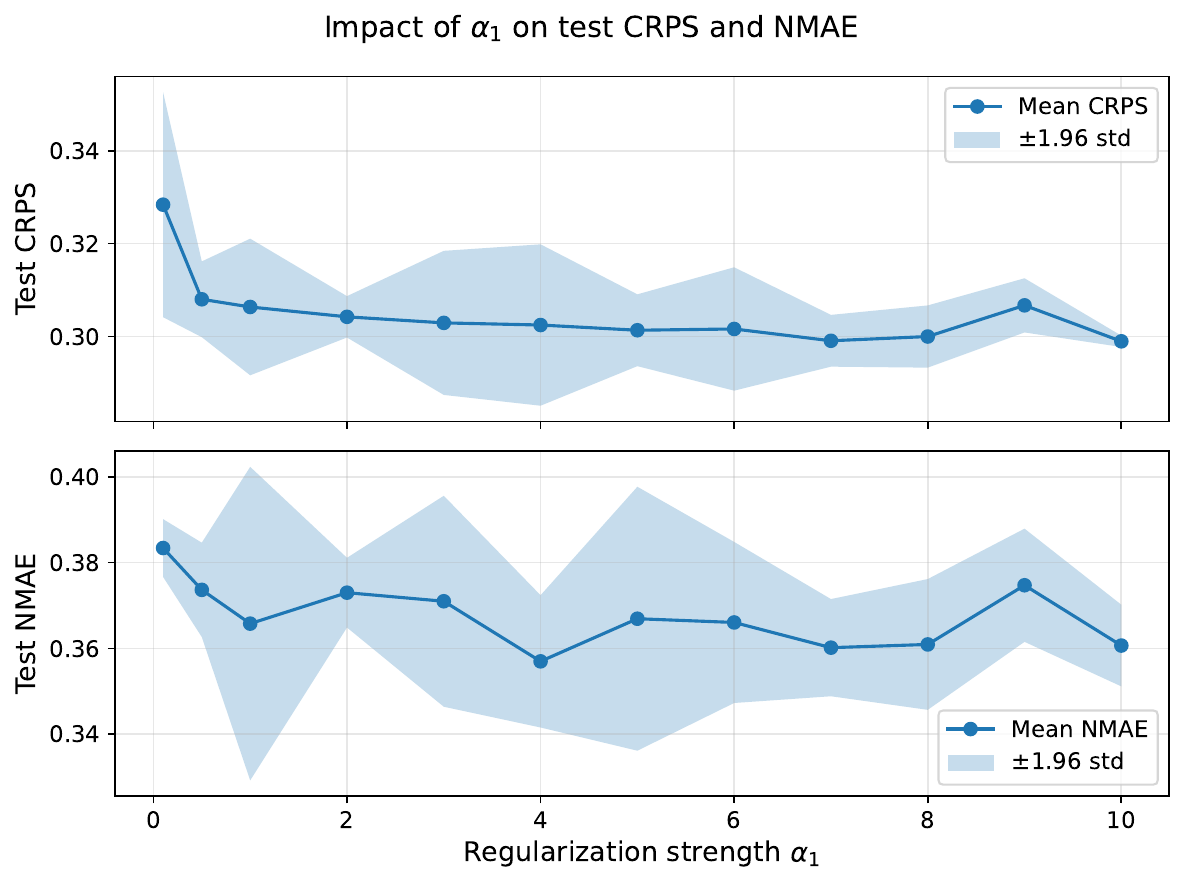}
    \caption{Effect of first-order latent trend regularization ($\alpha_1$) on long-term forecasting performance on ETTh1 ($T=96$, $L=720$).}
    \label{fig:alpha1_effect}
\end{figure}

\begin{figure}
    \centering
    \includegraphics[width=0.9\linewidth]{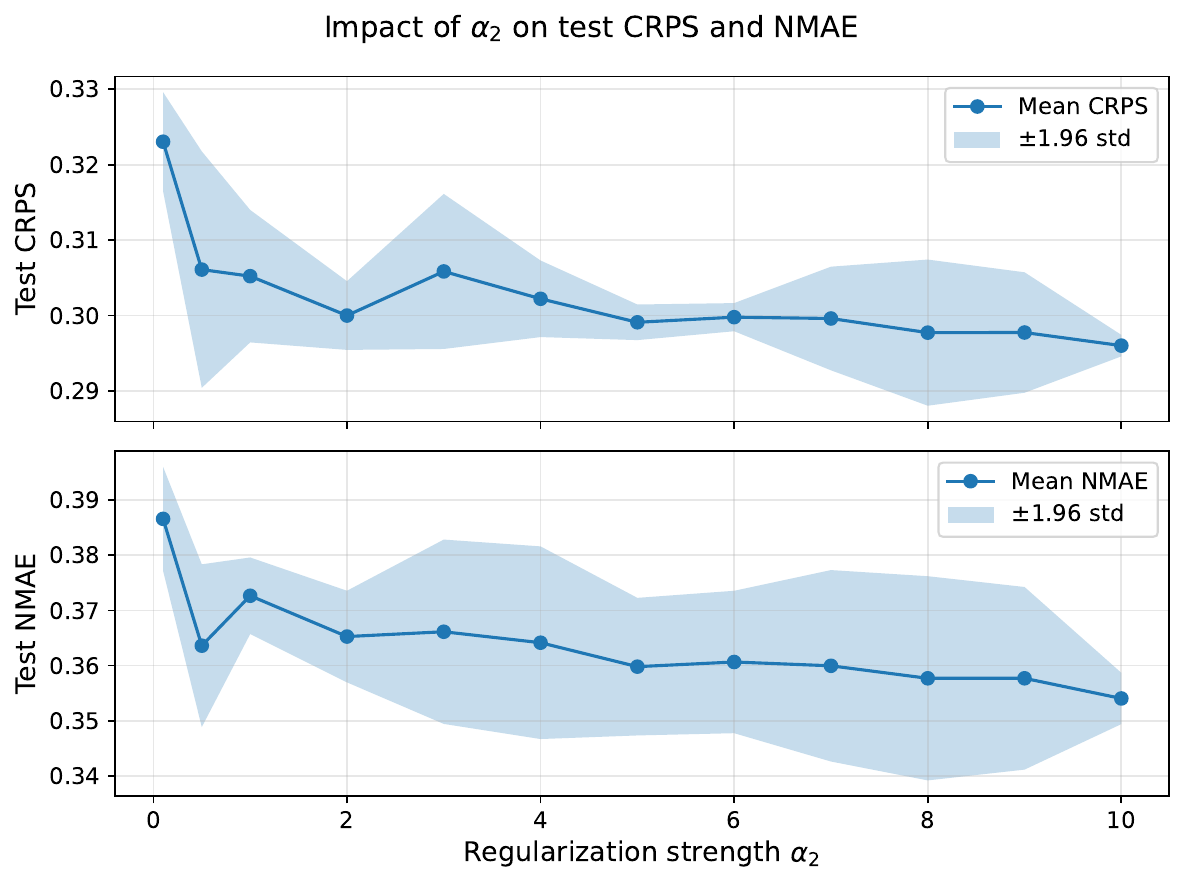}
    \caption{Effect of second-order latent trend regularization ($\alpha_2$) on long-term forecasting performance on ETTh1 ($T=96$, $L=720$).}
    \label{fig:alpha2_effect}
\end{figure}

To investigate the impact of our Hodrick-Prescott-inspired trend regularization, we conduct an ablation study varying the regularization strengths $\alpha_1$ and $\alpha_2$ on the ETTh1 dataset ($T = 96$, $L=720$). Figure~\ref{fig:alpha1_effect} and Figure~\ref{fig:alpha2_effect} demonstrate that both regularization terms improve forecasting performance.


Without first-order regularization ($\alpha_1 = 0$), the model achieves CRPS of $0.33$ and NMAE of $0.39$. Introducing moderate regularization ($\alpha_1 \in [0.5, 7.0]$) yields improvements of approximately $9\%$ in CRPS and $8\%$ in NMAE, stabilizing performance at CRPS $\approx 0.30$ and NMAE $\approx 0.36$. Similarly, second-order regularization provides consistent gains, with performance improving from the unregularized baseline when $\alpha_2 \in [1.0, 5.0]$. These results empirically validate Theorems~\ref{thm:hp_equivalent} and \ref{thm:nonlinear-decoder}, confirming that differential regularization successfully enforces trend smoothness and spectral separation, directly translating to improved forecasting accuracy.


\section{Conclusion}
We introduce DecoVAE, an efficient, interpretable variational autoencoder for probabilistic time series forecasting. By isolating trend and seasonal dynamics into distinct latent pathways, it applies domain-specific biases without interference: a time-domain differential regularizer ensures trend smoothness, while a complex Gaussian VAE captures seasonal cyclical behaviors in the frequency domain. Evaluated across seven real-world benchmarks, DecoVAE consistently outperforms strong baselines, improving CRPS and NMAE by up to 14.96\% and 23.30\% for short-term tasks, and 52.68\% and 26.51\% for long-term horizons. Crucially, it achieves these accuracy gains with substantial computational efficiency, reducing model weights by up to 93\% and accelerating runtime by 74\% versus the second-best baseline.



\bibliographystyle{IEEEtran}
\bibliography{references}

\appendix

\subsection{Dataset Specifications}
\label{app:datasets}
Our experimental study uses seven widely-adopted multivariate time series datasets with varying dimensionality, granularity, and domains~\cite{wu2025k2vae, zhang2024probts, zhou2021informer, wu2021autoformer}:
\begin{itemize}
\item \textbf{ETT:} Four variants (ETTh1/h2 hourly, ETTm1/m2 15-min) of transformer temperature metrics (2016--2018).
\item \textbf{Electricity:} Hourly consumption from 321 clients (2012--2014).
\item \textbf{Weather:} 10-min meteorological observations (temperature, humidity, pressure) throughout 2020.
\item \textbf{Traffic:} Hourly road occupancy from 862 San Francisco Bay Area sensors (2015--2016).
\end{itemize}
We augment each series with 11 time-dependent covariates: sinusoidal encodings for yearly to daily periodicities and a normalized time-delta feature. Following standard practice, we use chronological splitting ($70\%$ train, $10\%$ validation, $20\%$ test).

\begin{table*}[!htbp]
\centering
\footnotesize
\caption{Comparison of \textbf{different DecoVAE variants} on short-term forecasting task. ($T = 96$, $L=24$) Lower CRPS and NMAE scores indicate better performance. Results are reported as mean $\pm$ standard error over three independent runs.}
\label{tab:freq_variants}
\begin{tabular}{llccccc}
\toprule
Dataset & Metric & FreqMLP (w/o Reg.) & FreqMLP (w/ Reg.) & AmpPhase & FITS (w/o Reg.) & FITS (w/ Reg.) \\
\midrule

ETTh1
& CRPS & $0.229 \pm 0.002$ & $0.230 \pm 0.000$ & $0.234 \pm 0.002$ & \underline{$0.217 \pm 0.005$} & $\mathbf{0.215 \pm 0.001}$ \\
& NMAE & $0.265 \pm 0.002$ & $0.267 \pm 0.000$ & $0.268 \pm 0.002$ & \underline{$0.262 \pm 0.002$} & $\mathbf{0.259 \pm 0.001}$ \\
\midrule

ETTh2
& CRPS & $0.336 \pm 0.003$ & $0.345 \pm 0.005$ & $0.347 \pm 0.001$ & $\mathbf{0.319 \pm 0.005}$ & \underline{$0.327 \pm 0.014$} \\
& NMAE & $\mathbf{0.381 \pm 0.004}$ & $0.386 \pm 0.008$ & $0.386 \pm 0.000$ & \underline{$0.382 \pm 0.005$} & $0.383 \pm 0.000$ \\
\midrule

ETTm1
& CRPS & $\mathbf{0.184 \pm 0.002}$ & $0.194 \pm 0.001$ & $0.197 \pm 0.003$ & \underline{$0.188 \pm 0.002$} & $0.190 \pm 0.001$ \\
& NMAE & $\mathbf{0.217 \pm 0.001}$ & $0.229 \pm 0.002$ & $0.227 \pm 0.003$ & $0.225 \pm 0.002$ & \underline{$0.225 \pm 0.001$} \\
\midrule

ETTm2
& CRPS & $0.246 \pm 0.007$ & $0.247 \pm 0.001$ & $0.252 \pm 0.002$ & $\mathbf{0.233 \pm 0.001}$ & \underline{$0.234 \pm 0.001$} \\
& NMAE & $\mathbf{0.283 \pm 0.007}$ & $0.288 \pm 0.002$ & $0.286 \pm 0.002$ & \underline{$0.286 \pm 0.001$} & $0.287 \pm 0.003$ \\
\midrule

Weather
& CRPS & $0.297 \pm 0.015$ & $0.293 \pm 0.008$ & $0.304 \pm 0.004$ & \underline{$0.286 \pm 0.003$} & $\mathbf{0.278 \pm 0.006}$ \\
& NMAE & $0.330 \pm 0.014$ & $0.322 \pm 0.009$ & $0.324 \pm 0.004$ & \underline{$0.316 \pm 0.001$} & $\mathbf{0.305 \pm 0.007}$ \\

\bottomrule
\end{tabular}
\end{table*}

\subsection{Performance Metrics}
\label{app:metrics}
We quantify forecasting quality through two established metrics that assess complementary aspects of prediction performance:

\paragraph{Normalized Mean Absolute Error (NMAE).} This metric evaluates point forecast accuracy by normalizing absolute deviations relative to the true values:
\[
\mathrm{NMAE}(x, \hat{x}) = \frac{1}{N} \sum_{n=1}^N \frac{\sum_{i=1}^{F}\lvert x^n_i - \hat{x}^n_i \rvert}{\sum_{i=1}^{F}\lvert x^n_i \rvert},
\]
where $x^n_i$ represents the ground truth and $\hat{x}^n_i$ is the predicted median from 100 sampled trajectories.

\paragraph{Continuous Ranked Probability Score (CRPS).} To assess the quality of predictive distributions, we use CRPS, which compares the cumulative distribution function $F(z)$ against the empirical distribution of observations:
\[
\mathrm{CRPS} = \int_{\mathbb{R}} \bigl(F(z) - \mathbb{I}\{x \leq z\}\bigr)^2 dz.
\]
We compute this integral using Monte Carlo approximation following~\cite{zamo2018estimation}. Both metrics are negatively oriented, with smaller values indicating superior performance.

\subsection{Seasonal Module Variants}
\label{app:season_variants}

We compare real-valued, complex-valued, and amplitude-phase factorized spectral parameterizations for the seasonal component. To stabilize frequency-domain training, each variant employs tailored spectral regularization alongside standard reconstruction and KL terms.


\textbf{Complex-valued FITS-style model~\cite{xufits}.}
Seasonal embeddings are derived via complex-valued linear layers directly in the frequency domain, preserving the Fourier transform's structure by jointly capturing amplitude and phase. Consequently, spectral regularization enforces explicit energy matching between the input and latent representations. Let $E_x$ and $E_z$ denote the average energy of the input and latent signals, respectively:
\begin{align}
E_x &= \mathbb{E}\left[\|\mathbf{X}\|_2^2\right], \\
E_z &= \mathbb{E}\left[\|\mathbf{Z}_s\|_2^2\right].
\end{align}
The regularization term is defined as:
\begin{equation}
\mathcal{L}_{spec}^{\text{FITS}} = \left(E_x - E_z\right)^2 + \lambda \cdot \mathbb{E}\left[\|\mathbf{Z}_s\|_1\right],
\end{equation}
which jointly enforces energy preservation while penalizing noisy latent activations.

\textbf{Real-valued Frequency MLP (FreqMLP).}
Decomposing the complex spectrum into real and imaginary parts enables processing via standard real-valued networks, though this sacrifices explicit preservation of the Fourier domain's complex structure. Consequently, the real-valued frequency MLP applies a simple power spectral density (PSD) penalty to the latent representation:
\begin{equation}
\mathrm{PSD}(\mathbf{Z}_s) = \frac{|\mathbf{Z}_s|^2}{d}, \qquad
\mathcal{L}_{spec}^{\text{FreqMLP}} = \sum \mathrm{PSD}(\mathbf{Z}_s),
\end{equation}
which encourages bounded spectral energy and discourages excessive high-frequency amplification.

\textbf{Amplitude-Phase factorized model (AmpPhase).}
The Fourier spectrum is decomposed into amplitude and phase, modeled as independent stochastic variables via separate variational encoders for disentangled spectral modeling. Reconstruction occurs separately for magnitude and phase, with the amplitude loss integrating relative Frobenius error and log-magnitude consistency~\cite{yamamoto2020parallel}:
\begin{align}
\mathcal{L}_{amp} &= \frac{\|\widehat{A} - A\|_F}{\|A\|_F + \epsilon} + \sum \left|\log(\widehat{A} + \epsilon) - \log(A + \epsilon)\right|,
\end{align}
while the phase loss enforces angular alignment~\cite{barmpas2025advancing}:
\begin{equation}
\mathcal{L}_{phase} = \sum \left(1 - \cos(\phi - \widehat{\phi})\right).
\end{equation}
The total spectral loss is:
\begin{equation}
\mathcal{L}_{spec}^{\text{AmpPhase}} = \mathcal{L}_{amp} + \mathcal{L}_{phase}.
\end{equation}

Further implementation details of each variant are available in the accompanying GitHub repository\footnote{\url{https://anonymous.4open.science/r/DecoVAE-3387/}}. 


\subsubsection{Ablation Study on Spectral Parameterizations performance}
\label{app:seasonal_comparison}
An extended ablation study (Table~\ref{tab:freq_variants}) compares seasonal spectral parameterizations, including real-valued FreqMLP, amplitude-phase factorized, and complex-valued FITS-style formulations, both with and without regularization. The results validate the FITS-based variant as the default choice for the DecoVAE seasonal module.

\subsection{Comparison of decomposition methods}
\label{app:decomposition}

\begin{table}[!htbp]
\centering
\scriptsize
\caption{Comparison of \textbf{decomposition methods in DecoVAE-FITS} (T = 96, L = 720) on long-term forecasting task. Lower CRPS and NMAE scores indicate better performance. Results are reported as mean $\pm$ standard error over three independent runs. \textbf{Boldface} highlights the best-performing method, while \underline{underlining} denotes the second-best result.}
\label{tab:decomposition_methods}
\begin{tabular}{llccc}
\toprule
Dataset & Metric & MA & MoE & EMA \\
\midrule

ETTh1
& CRPS & $\mathbf{0.298 \pm 0.003}$ & \underline{$0.304 \pm 0.009$} & $0.356 \pm 0.003$ \\
& NMAE & \underline{$0.360 \pm 0.004$} & $\mathbf{0.357 \pm 0.003}$ & $0.393 \pm 0.002$ \\

\midrule
ETTh2
& CRPS & $\mathbf{0.498 \pm 0.006}$ & \underline{$0.510 \pm 0.006$} & $0.515 \pm 0.006$ \\
& NMAE & $0.585 \pm 0.002$ & \underline{$0.584 \pm 0.005$} & $\mathbf{0.584 \pm 0.004}$ \\

\midrule
ETTm1
& CRPS & $\mathbf{0.297 \pm 0.004}$ & \underline{$0.301 \pm 0.001$} & $0.305 \pm 0.001$ \\
& NMAE & \underline{$0.329 \pm 0.004$} & $\mathbf{0.327 \pm 0.002}$ & $0.330 \pm 0.001$ \\

\midrule
ETTm2
& CRPS & \underline{$0.468 \pm 0.005$} & $\mathbf{0.467 \pm 0.007}$ & $0.472 \pm 0.001$ \\
& NMAE & $\mathbf{0.524 \pm 0.001}$ & $0.536 \pm 0.008$ & \underline{$0.528 \pm 0.001$} \\

\bottomrule
\end{tabular}
\end{table}

\end{document}